\pdfoutput=1

\documentclass[11pt]{article}

\usepackage{acl}

\usepackage{times}
\usepackage{latexsym}

\usepackage[T1]{fontenc}

\usepackage[utf8]{inputenc}

\usepackage{microtype}

\usepackage{subfig}
\usepackage{amsmath}
\usepackage{amssymb}
\usepackage{graphicx}
\usepackage{makecell}
\usepackage{multirow}
\usepackage{longtable}
\usepackage{boldline}
\usepackage{hyperref}
\usepackage{footmisc}

\title{Schema Encoding for Transferable Dialogue State Tracking}

\author{Hyunmin Jeon \\
  Computer Science and Engineering \\
  POSTECH, Pohang, South Korea \\
  \texttt{jhm9507@postech.ac.kr} \\\And
  Gary Geunbae Lee \\
  Computer Science and Engineering \\
  Graduate School of Artificial Intelligence \\
  POSTECH, Pohang, South Korea \\
  \texttt{gblee@postech.ac.kr} \\}

\begin{document}
\maketitle
\begin{abstract}
Dialogue state tracking (DST) is an essential sub-task for task-oriented dialogue systems.
Recent work has focused on deep neural models for DST.
However, the neural models require a large dataset for training.
Furthermore, applying them to another domain needs a new dataset because the neural models are generally trained to imitate the given dataset.
In this paper, we propose \textbf{S}chema \textbf{E}ncoding for \textbf{T}ransferable \textbf{D}ialogue \textbf{S}tate \textbf{T}racking (SET-DST), which is a neural DST method for effective transfer to new domains.
Transferable DST could assist developments of dialogue systems even with few dataset on target domains.
We use a schema encoder not just to imitate the dataset but to comprehend the schema of the dataset.
We aim to transfer the model to new domains by encoding new schemas and using them for DST on multi-domain settings.
As a result, SET-DST improved the joint accuracy by 1.46 points on MultiWOZ 2.1.
\end{abstract}

\section{Introduction}
The objective of task-oriented dialogue systems is to help users achieve their goals by conversations.
Dialogue state tracking (DST) is the essential sub-task for the systems to perform the purpose.
Users may deliver the details of their goals to the systems during the conversations, e.g., what kind of food they want the restaurant to serve and at what price level they want to book the hotel.
Thus, the systems should exactly catch the details from utterances.
They should also communicate with other systems by using APIs to achieve users' goals, e.g., to search restaurants and to reserve hotels.
The goal of DST is not only to classify the users' intents but also to fill the details into predefined templates that are used to call APIs.

Recent work has used deep neural networks for DST with supervised learning.
They have improved the accuracy of DST; however, they require a large dataset for training.
Furthermore, they need a new dataset to be trained on another domain.
Unfortunately, the large dataset for training a DST model is not easy to be developed in real world.
The motivation of supervised learning is to make deep neural networks imitate humans.
But, they actually imitate the given datasets rather than humans.
Someones who have performed hotel reservation work could easily perform restaurant reservation work if some guidelines are provided, but neural models may have to be trained on a new dataset of the restaurant domain.
The difference between humans and neural models is that humans can learn how to read guidelines and to apply the guidelines to their work.
This is why transfer learning is important to train neural models on new domains.

In this paper, we propose \textbf{S}chema \textbf{E}ncoding for \textbf{T}ransferable \textbf{D}ialogue \textbf{S}tate \textbf{T}racking (SET-DST), which is a neural DST method with transfer learning by using dataset schemas as guidelines for DST.
The motivation of this study is that humans can learn not only how to do their work, but also how to apply the guidelines to the work.
We aim to make a neural model learn how to apply the schema guidelines to DST beyond how to fill predefined slots by simply imitating the dataset on multi-domain settings.
The schema includes metadata of the dataset, e.g., which domains the dataset covers and which slots have to be filled to achieve goals.
SET-DST has a schema encoder to represent the dataset schema, and it uses the schema representation to understand utterances and to fill slots.
Recently, transfer learning has been becoming important because development of new datasets is costly.
Transfer learning makes it possible to pre-train neural models on large-scale datasets to effectively fine-tune the models on small-scale downstream tasks.

\begin{figure*}
    \centering
    \subfloat[Schema encoding for active slots and intents classification.]{
    \label{subfig:overview1}
	\includegraphics[width=0.4\textwidth]{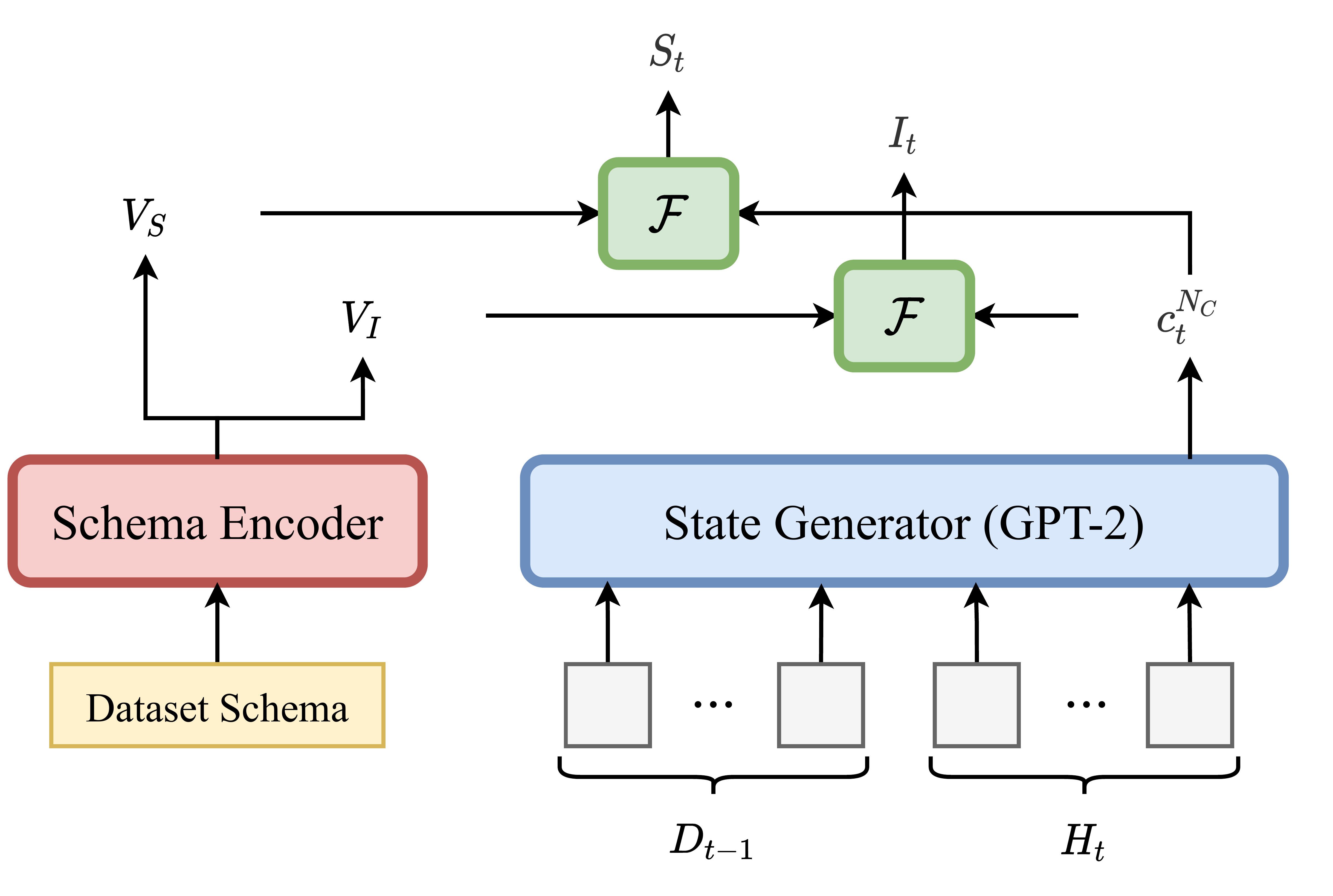}}
	\:\:\vrule\:\:
	\subfloat[Dialogue state generation.]{
    \label{subfig:overview2}
	\includegraphics[width=0.5\textwidth]{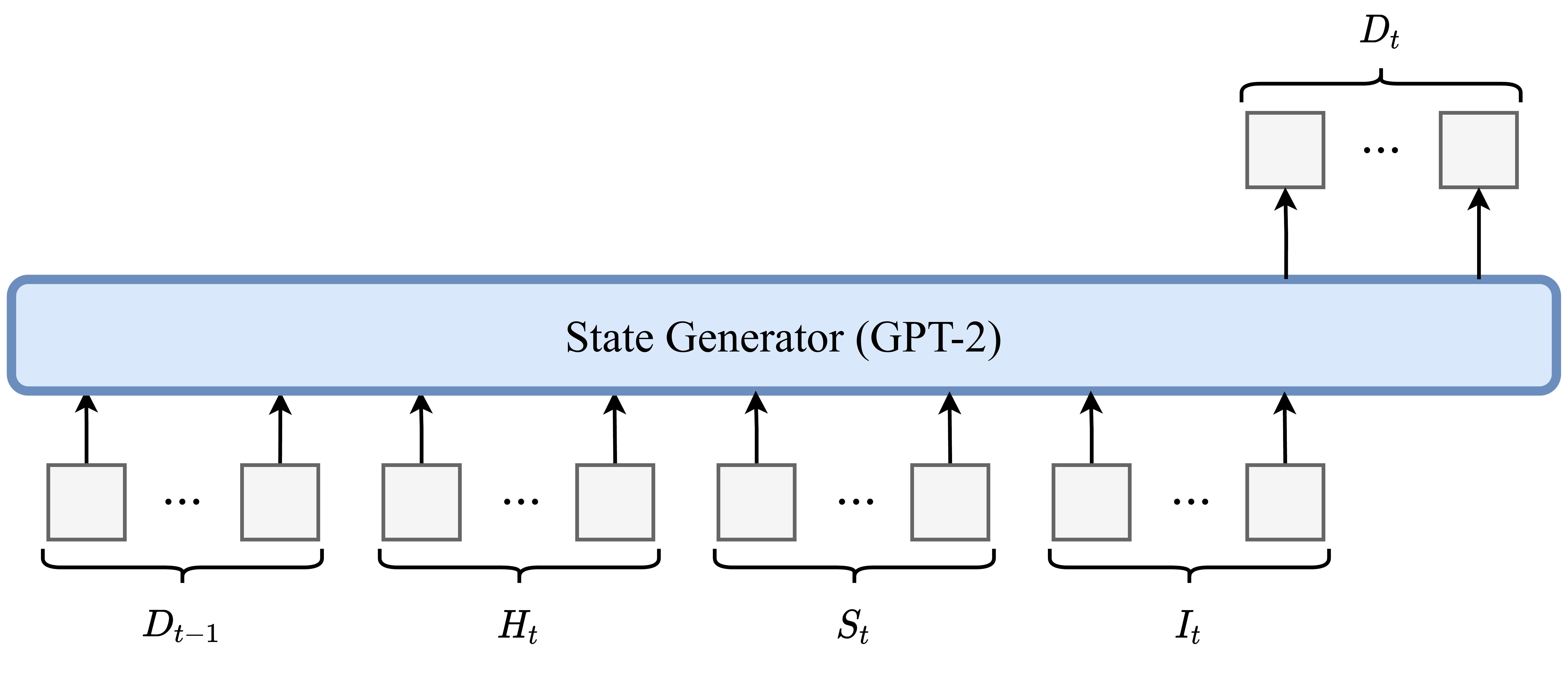}}
	\caption{Overview of SET-DST.
	The schema encoder takes the dataset schema and generates slot vectors and intent vectors.
	The state generator takes the previous dialogue state $D_{t-1}$ and the dialogue history $H_t$ to calculate active scores of slots and intents.
	$\mathcal{F}$ is an score function to calculate whether the slots or intents are activated on turn $t$.
	Then, the state generator additionally takes the activated slots and intents to generate the current dialogue state $D_t$.
	$S_t$ indicates the activated slots and $I_t$ indicates the activated intents.}
    \label{fig:overview}
\end{figure*}

We used SGD \citep{rastogi2020towards} as the large-scale dataset, and evaluated SET-DST on MultiWOZ 2.1 \citep{eric2020multiwoz}, which is a standard benchmark dataset for DST, as the downstream task.
SET-DST achieved state-of-the-art accuracy on the downstream DST task.
We further confirmed that SET-DST worked well on the small downstream dataset.
This result demonstrates that transfer learning with schema encoding improves the performance of neural DST models and the efficiency of few-shot learning on DST.

\section{Related Work}
Traditional DST models extract semantics by using natural language understanding (NLU) modules to generate dialogue states \cite{williams2014web, wang2013simple}.
The limitation of these models is that they rely on features extracted by humans.

Recent work has focused on building end-to-end DST models without hand-crafted features.
\citet{zhong2018global} use global modules to share parameters between different slots.
\citet{nouri2018toward} improve the latency by removing inefficient recurrent layers.
Transferable DST models that can be adapted to new domains by removing the dependency on the domain ontology are proposed \cite{ren2018towards, wu2019transferable}.
\citet{zhou2019multi} attempt to solve DST as a question answering task using knowledge graph.

More recently, large-scale pre-trained language models such as BERT \cite{devlin2019bert} and GPT-2 \cite{radfordlanguage} are used for DST. The pre-trained BERT acts as an NLU module to understand utterances \cite{lee2019sumbt, zhang2020find, kim2020efficient, heck2020trippy}.
GPT-2 makes it possible to solve DST as a conditional language modeling task \cite{hosseini2020simple, peng2021soloist}.

\citet{rastogi2020towards} propose the baseline method that defines the schema of dataset and uses it for training and inference.
A drawback of them is that the calculation cost is high because they use the domain ontology and access all values to estimate the dialogue state.
DST models that uses schema graphs to encode the relation between slots and values are proposed \cite{chen2020schema, zhu2020efficient}.
However, they focus on encoding the relation between slots and values of the given domains not on adaptation to new domains.

In this paper, we focus on making the model learn how to understand the schema and how to apply it to estimate the dialogue state, not just on encoding the in-domain relation.

\section{Schema Encoding for Transferable Dialogue State Tracking}
In this section, we describe the architecture of SET-DST and how to optimize it.
Figure \ref{fig:overview} shows the overview of our method.
The model consists of the schema encoder and the state generator.
SET-DST generates the dialogue state in two steps: (a) schema encoding and classification, and (b) dialogue state generation.
In this paper, we define some terms as follows.
\paragraph{Schema}
Metadata of the dataset, e.g., what domains, services, slots, and intents the dataset covers.
A dataset has a schema that describes the dataset.
\paragraph{Domain}
What domains the conversation goes on, e.g., restaurant, hotel, and attraction.
A conversation can go on multiple domains.
\paragraph{Service}
What services the system provides to users. 
It is similar to domain, but application-level.
For example, restaurant domain can have two different services: (1) a service for searching and reserving restaurants and (2) a service focused on searching and comparing restaurants.
In real world, a service corresponds to an application.
\paragraph{Action}
Abstract actions of users to achieve their goals during conversations, e.g., to inform the system their requirements or to request the system for some information.
Appendix \ref{sec:appendix:action} demonstrates the details of the user actions covered in this paper.
\paragraph{Slot}
The details of the user goals, e.g., the type of food and the price range of hotel.
Slots are predefined based on the domains or services that the system should cover, and the slots are filled by DST.
The schema includes the information of slots.
\paragraph{Value}
The values that have actual meaning for the corresponding slots, e.g., cheap or expensive about the price range of hotel.
The systems should match slot-value pairs from conversations.
\paragraph{Intent}
Sub-goals to achieve the final goals of users.
A goal consists of one or more intents, and an intent is achieved over one or more conversation turns.
In real world, an intent corresponds to an API.
For example, to search restaurants or to book hotels should be performed by APIs of external systems.
Furthermore, The dialogue system should predict the slot-value pairs which correspond to arguments to call APIs.

\begin{figure}
    \centering
	\includegraphics[width=0.45\textwidth]{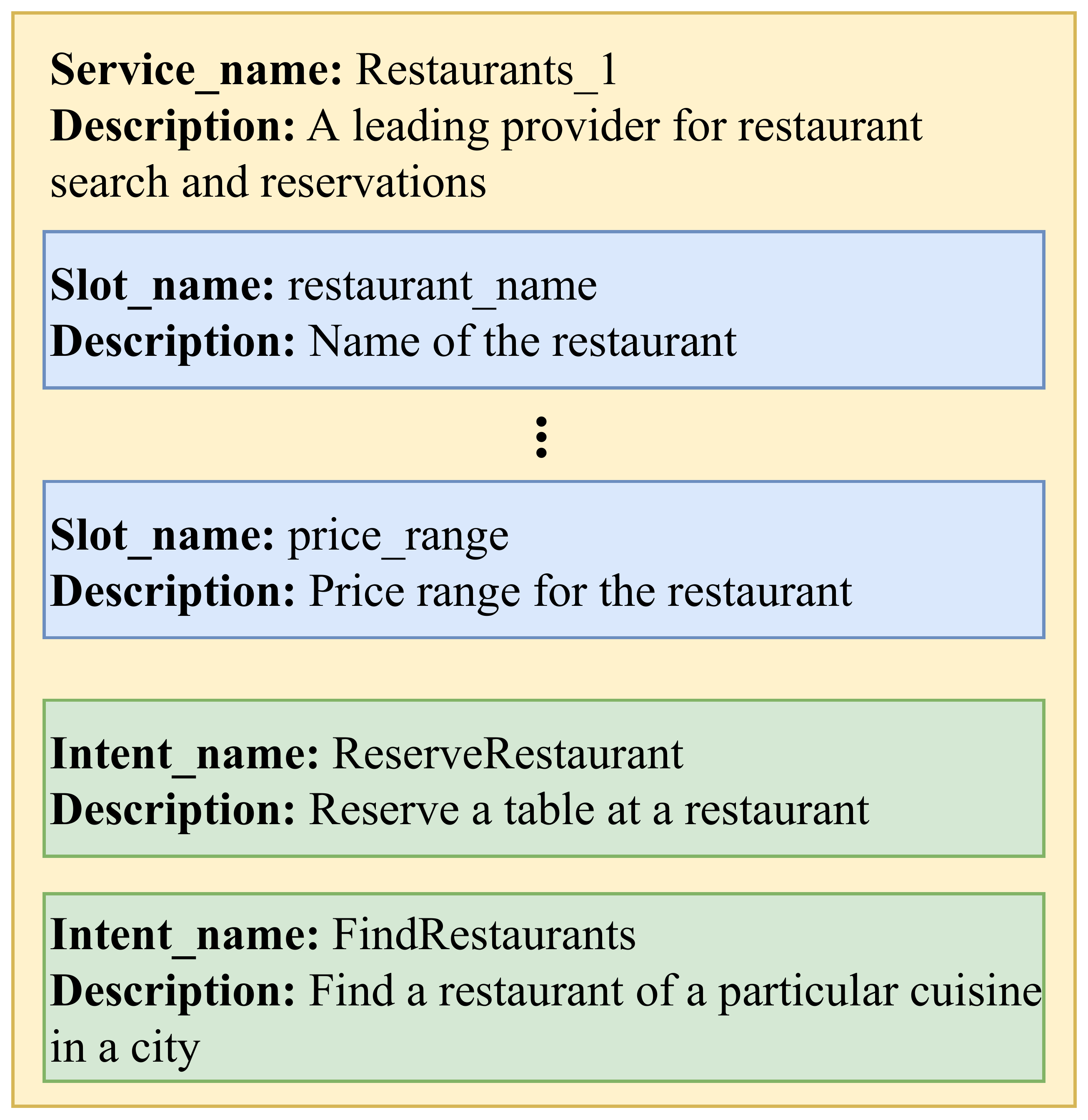}
	\caption{Example of schema for restaurant search and reservation service including slots and intents.}
    \label{fig:schema}
\end{figure}

\subsection{Schema Encoding}
We use the pre-trained BERT\footnote{\label{note1}The pre-trained models are available at \url{https://github.com/huggingface/transformers}.} for the schema encoder.
Figure \ref{fig:schema} shows an example of the schema for \textit{Restaurant\_1} service that is a service to search and reserve restaurants.
Services, slots, and intents consist of name and short description.
The name and description of the service in the schema are fed into BERT to generate service vector $v_R$ as
\begin{align}
    \label{eq1}
    \begin{split}
        o_R &= \texttt{BERT} \left( \texttt{[CLS]} n_R : d_R \texttt{[SEP]} \right) \\
        v_R &= W_R \cdot o_R^{\texttt{[CLS]}} \in \mathbb{R}^h
    \end{split},
\end{align}
where $n_R$ is the service name, $d_R$ is the service description, and $h$ is the hidden size.
$o_R^{\texttt{[CLS]}}$ is the output of \texttt{[CLS]} token, and $W_R \in \mathbb{R}^{h \times h}$ is a fully connected (FC) layer.
$\texttt{[CLS]}$ and $\texttt{[SEP]}$ are special tokens that mean the start and end of the sentence, respectively.
The service in Figure \ref{fig:schema} can be represented as \texttt{[CLS] Restaurants\_1 : A leading provider for restaurant search and reservations [SEP]} to be fed into BERT.
The slots and intents in the schema are also fed into BERT to generate slot vectors $V_S = \{v_S^1, \cdots v_S^{N_S}\} \in \mathbb{R}^{N_S \times h}$ and intent vectors $V_I = \{v_I^1, \cdots, v_I^{N_I}\} \in \mathbb{R}^{N_I \times h}$, respectively, as follows:
\begin{align}
    \label{eq2}
    \begin{split}
        o_S^j &= \texttt{BERT} \left( \texttt{[CLS]} n_S^j : d_S^j \texttt{[SEP]} \right) \\
        v_S^j &= W_S \cdot o_S^{j, \texttt{[CLS]}} \in \mathbb{R}^h, \quad j \in [1, N_S]
    \end{split}, \\
    \label{eq3}
    \begin{split}
        o_I^k &= \texttt{BERT} \left( \texttt{[CLS]} n_I^k : d_I^k \texttt{[SEP]} \right) \\
        v_I^k &= W_I \cdot o_I^{k, \texttt{[CLS]}} \in \mathbb{R}^h, \quad k\in[1, N_I] 
    \end{split}.
\end{align}
$N_S$ and $N_I$ mean the number of slots and intents for the service, respectively.
$n_S^j$ is the $j$-th slot name, and $d_S^j$ is the $j$-th slot description.
$o_S^{j, \texttt{[CLS]}}$ is the output of \texttt{[CLS]} token from the $j$-th slot, and $W_S \in \mathbb{R}^{h \times h}$ is an FC layer.
Similarly, $n_I^k$ is the $k$-th intent name, and $d_I^k$ is the $k$-th intent description.
$o_I^{k, \texttt{[CLS]}}$ is the output of \texttt{[CLS]} token from the $k$-th intent, and $W_I \in \mathbb{R}^{h \times h}$ is an FC layer.
The schema encoder takes $v_R$, $V_S$, and $V_I$ to update the slot vectors $V_S$ and intent vectors $V_I$ with attention mechanism as follows:
\begin{align}
    \label{eq4}
    \begin{split}
        a_S &= \texttt{softmax} \left( V_S \cdot v_R \right) \in \mathbb{R}^{N_S} \\
        v_{R, S} &= \left( V_S \right)^T \cdot a_S \in \mathbb{R}^h
    \end{split}, \\
    \label{eq5} v_S^j &= W_{RS} \cdot \left( v_{R, S} \oplus v_S^j \right) \in \mathbb{R}^h, \\
    \label{eq6}
    \begin{split}
        a_I &= \texttt{softmax} \left( V_I \cdot v_R \right) \in \mathbb{R}^{N_I} \\
        v_{R, I} &= \left( V_I \right)^T \cdot a_I \in \mathbb{R}^h
    \end{split}, \\
    \label{eq7} v_I^k &= W_{RI} \cdot \left( v_{R, I} \oplus v_I^k \right) \in \mathbb{R}^h.
\end{align}
$W_{RS} \in \mathbb{R}^{h \times 2h}$ and $W_{RI} \in \mathbb{R}^{h \times 2h}$ are FC layers, and $\oplus$ means the concatenation of two vectors.
The slot vectors and intent vectors updated with reference to the service vector are used for next steps: classification and generation.

\subsection{Slot and Intent Classification}
SET-DST takes the slot vectors and intent vectors to classify what slots and intents are activated by users.
We use the pre-trained GPT-2\footref{note1} for the state generator that encodes the dialogue history and generates the dialogue state as a sequence of words.
The state generator encodes the dialogue history $H_t$ that is accumulated during the conversation and the previous dialogue state $D_{t-1}$ to calculate the context vector $C_t $ as
\begin{align}
    \label{eq8} \begin{split}
        C_t &= \left\{ c_t^1, \cdots, c_t^{N_C} \right\} \in \mathbb{R}^{N_C \times h} \\
        &= \texttt{GPT-2} \left( D_{t-1} \oplus H_t \right),
    \end{split}
\end{align}
where $N_C = |C_t|$, and $c_t^i$ means the GPT-2 output of the $i$-th word.
Then, the last output of $C_t$ is used to classify which slots and intents are activated in the current conversation as follows:
\begin{align}
    \label{eq9} P \left( s_t^j=\texttt{Active} \right) &= \mathcal{F} \left( c_t^{N_C}, v_S^j \right), \\
    \label{eq10} P \left( i_t^k=\texttt{Active} \right) &= \mathcal{F} \left( c_t^{N_C}, v_I^k \right),
\end{align}
where $P(s_t^j=\texttt{Active})$ means the probability that the $j$-th slot is activated on turn $t$, and $P(i_t^k=\texttt{Active})$ means the probability that the $k$-th intent is activated on turn $t$.
$v_S^j$ and $v_I^k$ indicate the slot vector of the $j$-th slot and the intent vector of the $k$-th intent, respectively, calculated by the schema encoder.
$\mathcal{F}$ is a projection layer to calculate the probabilities using the context vector, slot vector, and intent vector.
We define $\mathcal{F}(x,y)$ as a function transforming vectors $x$ and $y$ into a probability scalar as
\begin{align}
    \label{eq11}
    \begin{split}
        h_1 &= \texttt{tanh} \left( W_1 \cdot x \right) \\
        h_2 &= \texttt{tanh} \left( W_2 \cdot \left( h_1 \oplus y \right) \right) \\
        \mathcal{F}(x,y) &= \sigma \left( W_3 \cdot h_2 \right)
    \end{split},
\end{align}
where $W_1 \in \mathbb{R}^{h \times h}$, $W_2 \in \mathbb{R}^{h \times 2h}$, and $W_3 \in \mathbb{R}^{1 \times h}$ are FC layers.
Activate slots and intents are classified based on the probabilities $P(s_t^j=\texttt{Active})$ and $P(i_t^j=\texttt{Active})$.
We define the slots activated on turn $t$ as $S_t=\{s_t^j | P(s_t^j=\texttt{Active}) \geq \alpha, \: j\in[1, N_S]\}$ and the intents activated on turn $t$ as $I_t=\{i_t^k | P(i_t^k=\texttt{Active}) \geq \alpha, \: k\in[1, N_I]\}$.
$s_t^j$ and $i_t^k$ means the $j$-th slot name and the $k$-th intent name, respectively.
$\alpha$ is a threshold to classify the slots and intents based on the probabilities.
Activated slots $S_t$ and intents $I_t$ classified on this step are used to generate the dialogue state.
On the generation step, $S_t$ and $I_t$ further contains slot vectors and intent vectors, which are calculated on the encoding step, in addition to the names in the text form.
For example, $S_t = \{ \texttt{restaurant\_name, price\_range} \}$ can be represented as
\begin{align}
    \begin{split}
        E(``\texttt{Slots:\{restaurant\_name:}") \oplus \\
        v_S^1 \oplus E(``\texttt{;price\_range:}") \oplus v_S^2 \oplus E(``\texttt{\}}")
    \end{split}
\end{align}
before being fed into GPT-2, where $E$ is the embedding layer to project the slot names into vector space of the same size as the slot vectors.
To generate the dialogue state that consists of the slot-value pairs, the system should recognize not only the values but also the name of slots.
The values can be extracted from user utterances, and which slots are activated can be predicted by the classification step; however, the exact slot names should be provided in the text form to construct slot-value structure matching given schema.
By this process, the system can recognize the name of activated slots and intents before generating the dialogue state.

\begin{figure}[ht]
    \centering
	\subfloat[Example in restaurants domain.]{
    \label{subfig:state:example1}
	\includegraphics[width=0.48\textwidth]{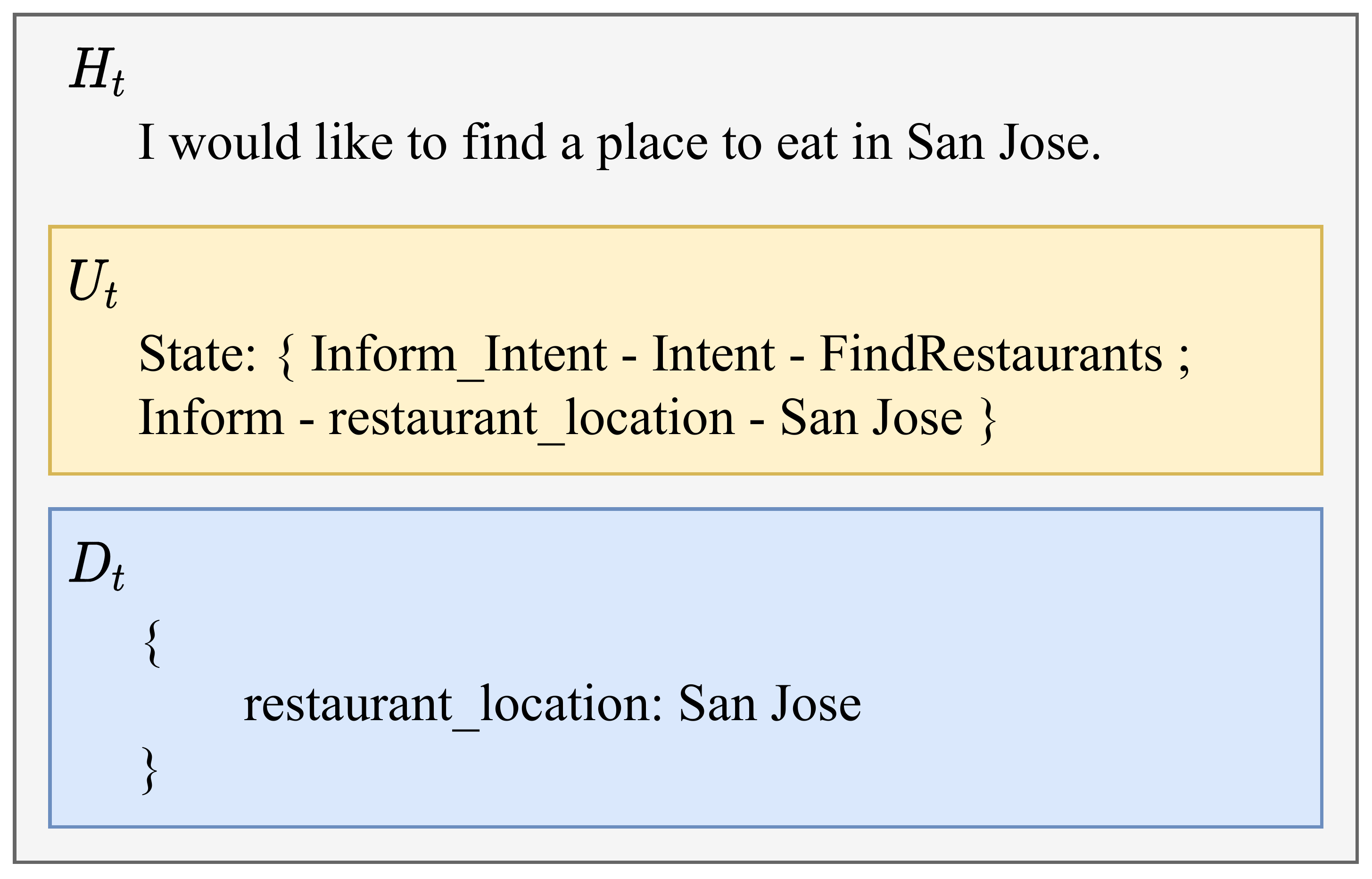}} \\
	\subfloat[Example in flights domain.]{
    \label{subfig:state:example2}
	\includegraphics[width=0.48\textwidth]{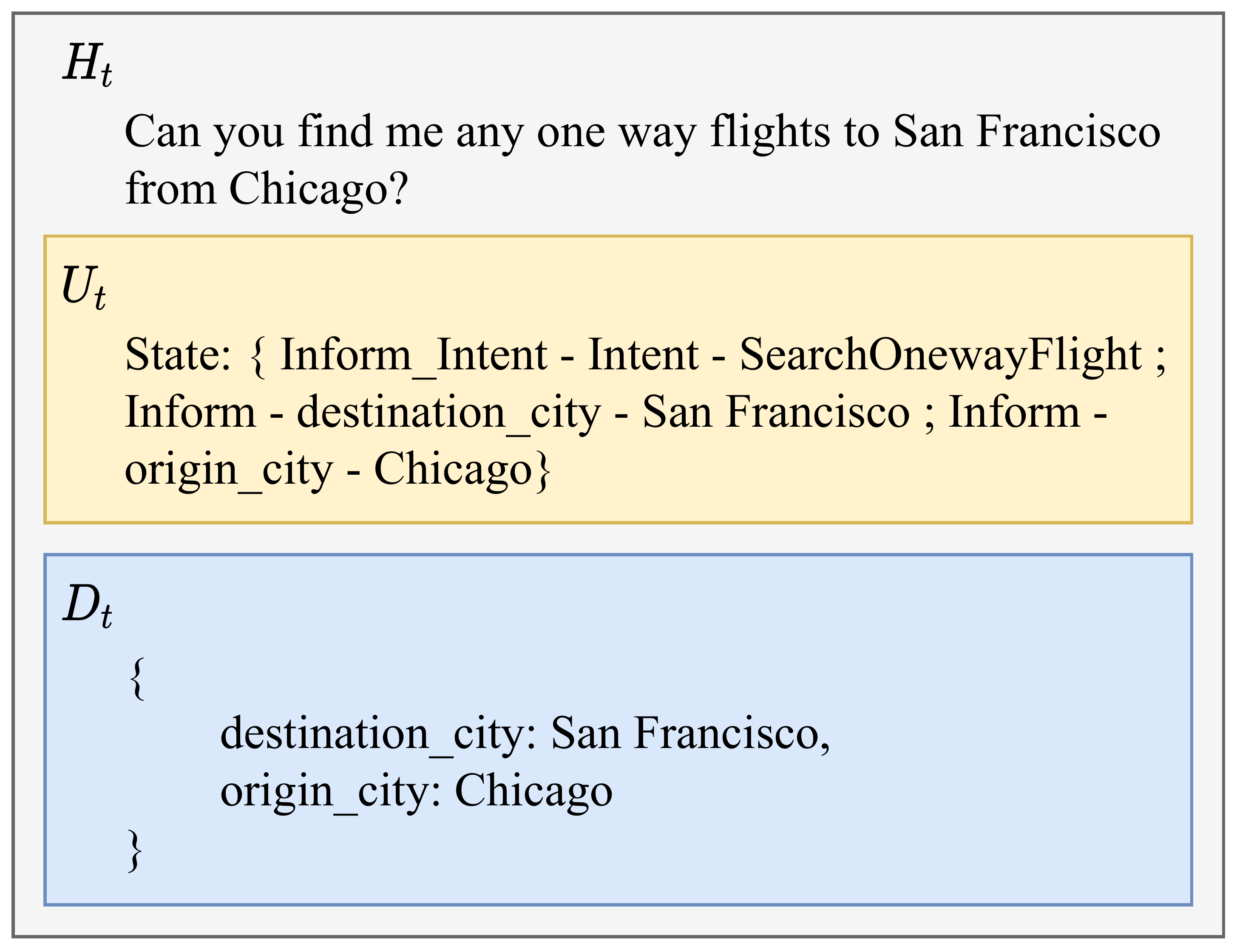}}
	\caption{Examples of user state and dialogue state corresponding to user utterance. $U_t$ is a sequence of words, and $D_t$ is a list of slot-value pairs.}
    \label{fig:state:example}
\end{figure}

\subsection{Dialogue State Generation}
SET-DST has the state generator that generates dialogue state using the dialogue history, schema representation, and previous dialogue state accumulated during the conversation.
In this paper, we define the dialogue state as a list of slot-value pairs that mean the details of an user goal.
We also define the concept called user state that is a sequence of action-slot-value triples to generalize semantics from various user utterances.
The state generator recurrently generates the user state as a sequence of words, instead of generating the dialogue state in the structured form directly.
Then, the dialogue state is updated by extracting the slot-value pairs from the user state.
The user state $U_t$ on turn $t$ is generated based on the previous dialogue state $D_{t-1}$, dialogue history $H_t$, active slots $S_t$, and active intents $I_t$ as follows:
\begin{align}
    \label{eq12}
    \tilde{u}_t^l &= \texttt{GPT-2} \left( D_{t-1} \oplus H_t \oplus S_t \oplus I_t \oplus U_t^{1:l-1} \right ), \\
    \label{eq13}
    \begin{split}
        U_t &= \Big\{ u_t^l \Big| u_t^l = \texttt{argmax} \left( W_{vocab} \cdot \tilde{u}_t^l \right), \\
        & \quad\quad l \in [1, N_U] \Big\} \in \mathbb{R}^{N_U},
    \end{split}
\end{align}
where $U_t^{1:l-1} = \{u_t^1, \cdots, u_t^{l-1}\}$ and $N_U = |U_t|$.
$u_t^l$ means the $l$-th word of the user state, and $W_{vocab} \in \mathbb{R}^{N_{vocab} \times h}$ is an FC layer to project the hidden state to vocabulary space with size of $N_{vocab}$.
Figure \ref{fig:state:example} shows how to generate $D_t$ from $U_t$.
$U_t$ is generated word-by-word over time steps until \texttt{[EOS]}, a special word to terminate the generation, is detected.
Then, $D_t$ is updated by extracting the slot-value pairs from $U_t$.
In task-oriented dialogue system, the dialogue state is used to call API.
In Figure \ref{subfig:state:example1}, \texttt{San Jose} is passed as the value of an argument named \texttt{restaurant\_location} to call the API named \texttt{FindRestaurants}.
However, we aim not to build the full task-oriented dialogue system but to generate the dialogue state in this study.

\subsection{Optimization}
SET-DST is optimized over two steps: (1) slot and intent classification, and (2) state generation.
We freeze the pre-trained BERT during training to preserve the broad and general knowledge that is learned from large corpus.
In classification task, the system is trained by using binary cross-entropy.
Equation \ref{eq9} is used to calculate the slot loss $\mathcal{L}_t^S$ with slot labels $Y_t^S = \{y_{t, 1}^S, \cdots, y_{t, N_S}^S\}$ as
\begin{align}
    \label{eq14}
    \begin{split}
        \mathcal{L}_t^S = &-\frac{1}{N_S}\sum_{j=1}^{N_S} \beta \cdot y_{t, j}^S \cdot \log P \left( s_t^j \right) \\
        &+ \left( 1 - y_{t, j}^S \right) \log \left( 1 - P \left( s_t^j \right) \right),
    \end{split}
\end{align}
where $y_{t, j}^S \in \mathbb{R}^1$ is the binary value of $j$-th slot on turn $t$, and $\beta$ is a hyperparameter to consider the ratio of active slots out of total slots.
Based on Equation \ref{eq10}, the intent loss $\mathcal{L}_t^I$ is calculated with intent labels $Y_t^I = \{y_{t, 1}^I, \cdots, y_{t, N_I}^I\}$ as
\begin{align}
    \label{eq15}
    \begin{split}
        \mathcal{L}_t^I = &-\frac{1}{N_I}\sum_{k=1}^{N_I} \beta \cdot y_{t, k}^I \cdot \log P \left( i_t^k \right) \\
        &+ \left( 1 - y_{t, k}^I \right) \log \left( 1 - P \left( i_t^k \right) \right),
    \end{split}
\end{align}
where $y_{t, k}^I \in \mathbb{R}^1$ is the binary value of $k$-th intent on turn $t$.
In state generation step, the system is trained as a conditional language model that recurrently generates words over time steps.
The state loss $\mathcal{L}_t^U$ is calculated base on Equation \ref{eq13} with the state label $Y_t^U = \{y_{t, l}^U, \cdots, y_{t, N_U}^U\}$ as 
\begin{align}
    \label{eq16}
    \mathcal{L}_t^U &= -\frac{1}{N_U} \sum_{l=1}^{N_U} \left( y_{t, l}^U \right)^T \log P \left( u_t^l \right),
\end{align}
where $y_{t, l}^U \in \mathbb{R}^{N_{vocab}}$ is the one-hot vector that indicates the $l$-th word of the gold-standard user state on turn $t$.
The final joint loss is the sum of above losses:
\begin{align}
    \label{eq17}
    \mathcal{L}_t &= \mathcal{L}_t^S + \mathcal{L}_t^I + \mathcal{L}_t^U.
\end{align}
We use Adam optimizer \cite{kingma2014adam} to minimize $\mathcal{L}_t$.

\section{Experiments}
In this section, we describe our experiments including the datasets, evaluation metric, and results.

\subsection{Experimental Setups}
We used two datasets MultiWOZ 2.1\footnote{\url{https://github.com/budzianowski/multiwoz}.} and Schema-Guided Dialogue (SGD)\footnote{\url{https://github.com/google-research-datasets/dstc8-schema-guided-dialogue}.} to evaluate our system.
MultiWOZ consists of conversations between a tourist and a guide, e.g., booking hotels and searching trains.
SGD deals with conversations between a virtual assistant and an user ranging over various domains, e.g., events, restaurants, and media.
The dataset also provides a schema that includes services, intents, and slots with short descriptions to help understanding the conversations.
In this study, we followed the schema proposed in SGD.
MultiWOZ has about 10,400 dialogues, and SGD has about 22,800 dialogues.

The datasets propose joint accuracy as the metric to evaluate DST systems.
Joint accuracy measures whether a system successfully predicts all slot-value pairs mentioned on the conversations.
In every turn, the system updates dialogue state, and the joint accuracy is calculated based on the accumulated dialogue state.

\begin{figure}
    \centering
	\includegraphics[width=0.45\textwidth]{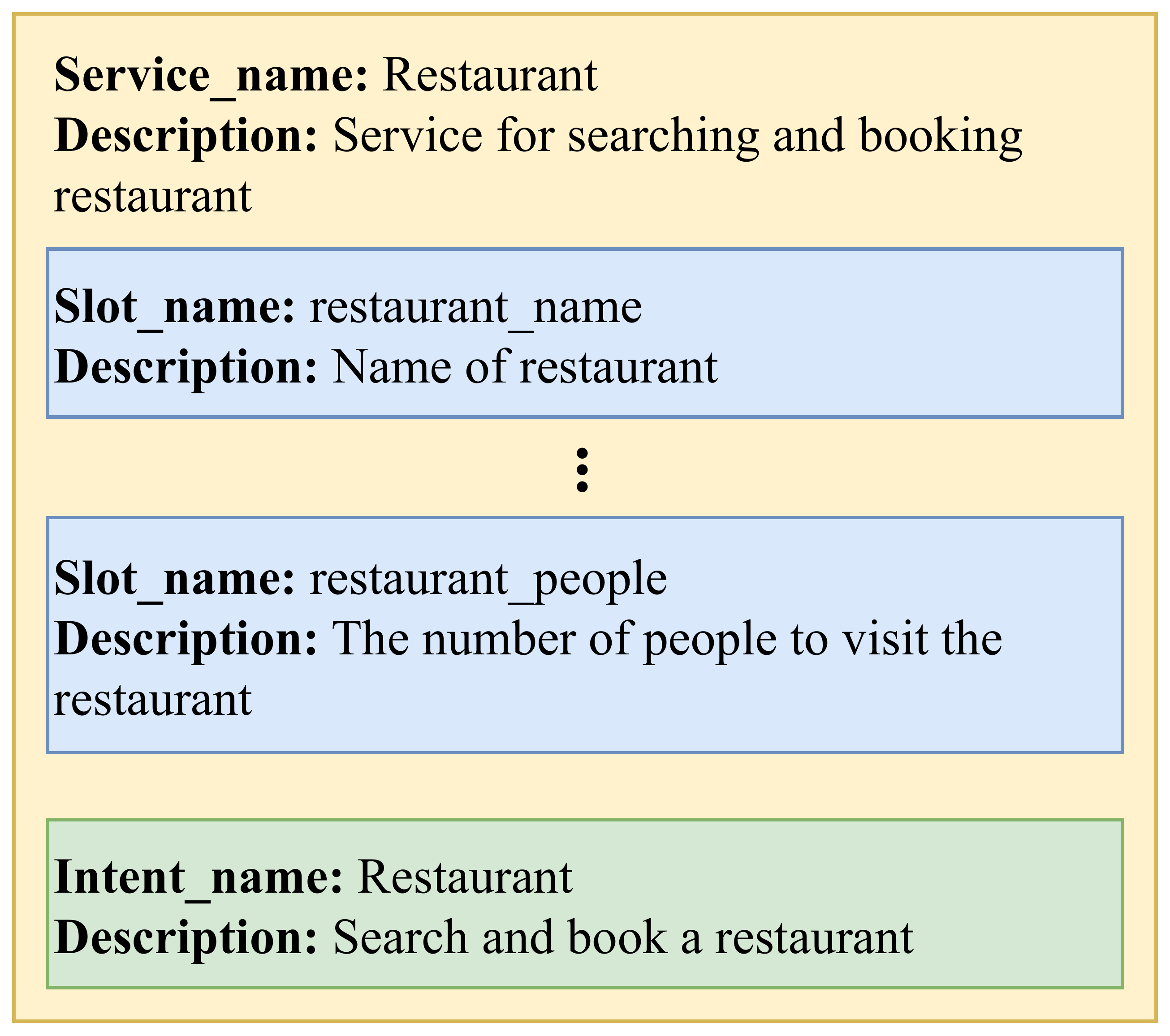}
	\caption{Example of schema that is temporarily created for MultiWOZ dataset.}
    \label{fig:schema:multiwoz}
\end{figure}
\begin{figure}[ht]
    \centering
	\includegraphics[width=0.45\textwidth]{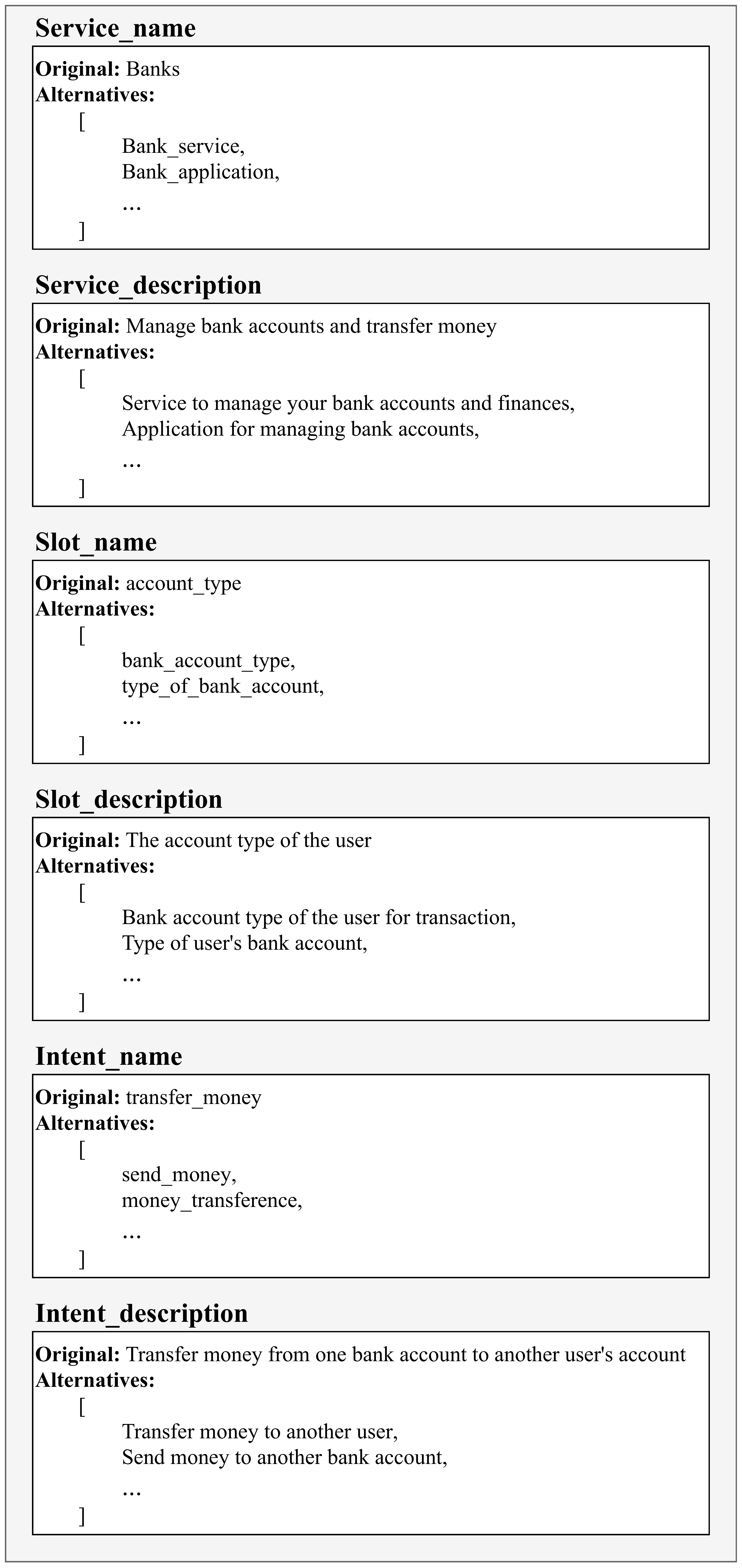}
	\caption{Example of alternatives for schema augmentation.}
    \label{fig:alter}
\end{figure}

\subsection{Experimental Details}
The motivation of SET-DST is to make the system interpret the schema and refer it for efficiently tracking the dialogue state.
In the experiments, our goal is to verify that SET-DST works well for our purpose by improving the performance of DST and the efficiency on few-shot settings with the schema encoding.

The experiments are divided into two steps: (1) pre-training on SGD and (2) fine-tuning on MultiWOZ.
In the pre-training step, SET-DST is optimized to encode the schema for DST.
In the fine-tuning step, the capability that encodes given schema is transferred to encode new schema for improvement of the performance and efficiency.
We conducted the experiments by adjusting the rate of few-shot data during fine-tuning to focus on the fine-tuning step.
The training data for few-shot settings was randomly sampled from the training set of MultiWOZ, and the random seed was fixed for consistency of sampling.
We also conducted experiments to verify whether SET-DST successfully works on the pre-training step, although the major part in our experiments is the fine-tuning on MultiWOZ including few-shot settings.

SET-DST needs not only slot information but also a schema.
However, MultiWOZ has no schema and no concepts of service and intent; thus, we created a schema for MultiWOZ including services, slots, intents, and corresponding descriptions.
Figure \ref{fig:schema:multiwoz} shows an example of the schema for MultiWOZ.
In our experiments on MultiWOZ, an intent means activated domain.
In other words, the system classifies an intent as active when the domain of conversation is changed or a conversation starts.
MultiWOZ further has no labels for activated intents, thus we automatically added the labels by tracking active domains and detecting whether new domains are active.

We further tried to fine-tune the system without intents because it is possible that the concepts of intent are unnatural in MultiWOZ.
In this setting, Equation \ref{eq3}, \ref{eq6}, \ref{eq7}, \ref{eq10}, \ref{eq15} are ignored, $I_t$ is removed from Equation \ref{eq12}, and $\mathcal{L}_t^I$ is removed from Equation \ref{eq17}.
In Figure \ref{subfig:state:example1}, $U_t$ is replaced with \texttt{State: \{ Inform - restaurant\_location - San Jose \}}, and in Figure \ref{subfig:state:example2}, $U_t$ is replaced with \texttt{State: \{ Inform - destination\_city - San Francisco ; Inform - origin\_city - Chicago \}}.

A dataset schema can be variously defined depending on the developer, and our goal is to make the system represent any schema for DST.
In the pre-training step, we manually augmented the schema of SGD dataset to avoid overfitting to the given schema.
The schema provides names of services, slots, and intents with short descriptions.
We defined some alternatives of the names and descriptions, and sampled inputs for the schema encoder from the augmented schema.
Figure \ref{fig:alter} shows some examples of the alternatives for bank service.

In dialogue state, multiple slots and intents can be activated at once.
However, the order has no meaning in dialogue state.
The state generator is trained to generate the dialogue state based on textual label, so it is possible that the order causes wrong optimization and overfitting.
Thus, we shuffled the order of slots and intents when making the labels.

We used BERT-base-uncased\footnote{\url{https://huggingface.co/bert-base-uncased}.} model for the schema encoder and GPT-2\footnote{\url{https://huggingface.co/gpt2}.} model for the state generator.
In our experiments, the pre-training step took about two days, and the fine-tuning step took about a day on a TitanRTX GPU.
Table \ref{table:hparams} lists hyperparameters that are used in our experiments.

\begin{table}
    \renewcommand{\tabcolsep}{5pt}
    \centering
    \begin{tabular}{l|l}
        \Xhline{5\arrayrulewidth} & \multicolumn{1}{c}{JA} \\
        \hline TRADE \cite{wu2019transferable} & 45.60\%* \\
        DSTQA \cite{zhou2019multi} & 51.17\% \\
        LABES-S2S \cite{zhang2020probabilistic} & 51.45\% \\
        DST-Picklist \cite{zhang2020find} & 53.30\% \\
        MinTL-BART \cite{lin2020mintl} & 53.62\% \\
        TripPy \cite{heck2020trippy} & 55.29\% \\
        SimpleTOD \small\cite{hosseini2020simple} & 55.76\% \\
        PPTOD \cite{su2021multi} & 57.45\% \\
        ConvBERT-DG \cite{mehri2020dialoglue} & 58.70\% \\
        TripPy+SCoRe \cite{yu2020score} & 60.48\% \\
        TripPy+CoCoAug \cite{li2020coco} & 60.53\% \\
        TripPy+SaCLog \cite{dai2021preview} & 60.61\% \\
        \hline SET-DST (Ours) & 60.39\% \\
        SET-DST w/o intent & \textbf{62.07}\% \\
        \Xhline{5\arrayrulewidth}
    \end{tabular}
    \caption{DST results on the test set of MultiWOZ in joint accuracy. *: the result is reported by \citet{eric2020multiwoz}.}
    \label{table:result}
\end{table}

\subsection{Experimental Results}
Table \ref{table:result} compares the evaluation results of SET-DST to the previous methods on the test set of MultiWOZ.
In our experiments, SET-DST achieved new state-of-the-art joint accuracy when fine-tuned without intent.

Table \ref{table:fewshot} shows the evaluation results on few-shot settings and the improvement by pre-training.
When we used less training data, the pre-training with schema encoding was more effective for DST.
SET-DST performed reasonably well with only about 20\% of the training data.
In most cases, the models fine-tuned without intents achieved higher joint accuracy on MultiWOZ.

\begin{table}
    \renewcommand{\tabcolsep}{5pt}
    \centering
    \begin{tabular}{cc|c|c}
        \Xhline{5\arrayrulewidth} \multicolumn{2}{c|}{\multirow{2}{*}{Few-shot rate}} & \multicolumn{2}{c}{JA} \\
        \cline{3-4} & & w/ intent & w/o intent \\
        \hline \multicolumn{1}{c|}{} & 100\% & 60.39\% & 62.07\% \\
        \multicolumn{1}{c|}{} & 30\% & 53.43\% & 56.43\% \\
        \multicolumn{1}{c|}{w/} & 25\% & 53.07\% & 55.61\% \\
        \multicolumn{1}{c|}{pre-training} & 20\% & 52.73\% & 54.37\% \\
        \multicolumn{1}{c|}{} & 15\% & 40.41\% & 51.29\% \\
        \multicolumn{1}{c|}{} & 10\% & 31.20\% & 29.91\% \\
        \hline \multicolumn{1}{c|}{} & 100\% & 58.37\% & 59.10\% \\
        \multicolumn{1}{c|}{} & 30\% & 31.08\% & 48.96\% \\
        \multicolumn{1}{c|}{w/o} & 25\% & 30.39\% & 30.28\% \\
        \multicolumn{1}{c|}{pre-training} & 20\% & 22.80\% & 22.35\% \\
        \multicolumn{1}{c|}{} & 15\% & 19.38\% & 17.09\% \\
        \multicolumn{1}{c|}{} & 10\% & 10.21\% & 15.10\% \\
        \Xhline{5\arrayrulewidth}
    \end{tabular}
    \caption{Evaluation results on few-shot settings with considering pre-training.}
    \label{table:fewshot}
\end{table}

\section{Discussion}
\paragraph{Schema Encoding}
In this study, our goal is to transfer a pre-trained DST model to a low-resource domain without limiting the transference as language model level by using schema encoding.
We pre-trained SET-DST on SGD which is a relatively large dataset and fine-tuned it on MultiWOZ to transfer the schema encoding.
As a result, the pre-training significantly improved the accuracy on DST.
Table \ref{table:fewshot} shows that the pre-training was more effective when the target dataset was small.
To satisfy the joint accuracy evaluation, the system should not only predict values, but also exactly match the names of slots defined on the dataset.
We believe that the tuning process was not completed when we used just 10\% of the target dataset; on the other side, the system could match the slot names defined on MultiWOZ not SGD with 20\% of the dataset.
Pre-trained language models have been already used in many fields.
However, our method could tackle general DST beyond language modeling on various domains.
We believe that SET-DST can assist the development of DST systems in real world without large dataset on the target domain.

\paragraph{Intent on Fine-tuning}
In this paper, we define the intents as sub-goals to be achieved through a service.
SGD has a schema for dialogues between a virtual assistant and an user.
Thus, it is assumed that a system achieves the user's sub-goals by using APIs, and an intent corresponds to an API.
Virtual assistant should tackle various services that could consist of one more intents, e.g., to check account balance and to transfer money in bank service.
Unlike that, MultiWOZ has no schema and considers no APIs as intents.
Thus, the schema that we temporarily created for experiments in the same form as the schema of SGD could cause confusion in generation of dialogue state.
We believe that this is why the results without intent were slightly higher in the experiments.
Another reason would be the incorrect labels for intents that we automatically created for experiments on MultiWOZ.

The joint accuracy that has been proposed as an evaluation metric for DST considers only slot-value pairs.
However, task-oriented dialogue systems should call APIs of external systems to achieve goals, e.g., to search restaurants and to reserve hotels.
The systems that predict only slot-value pairs would be insufficient to replace rule-based traditional systems in real-world.
Even though use of intents made no improvement in joint accuracy, we believe that encoding the schema including intents is meaningful in terms of approaching more realistic DST.

\paragraph{Backbone Model}
We used BERT for schema encoding and GPT-2 for classification and generation.
Using only GPT-2 was an option in our study.
However, we fixed the pre-trained BERT during optimization to preserve the broad and general knowledge learned on pre-training and to use the knowledge for encoding given schema.
BERT is an excellent backbone model to encode text for downstream tasks; thus, we used BERT instead of GPT-2 for the schema encoding.

\section{Conclusion}
Transfer learning that makes it possible to apply a pre-trained model to new domains has been attempted a lot.
However, the attempts for DST have been just to use large-scale pre-trained models as language models.
In this paper, we have proposed SET-DST, which is an effective method for DST with transfer learning by using schema encoding.
We have demonstrated how to encode the schema for transferable DST and how to use the schema representation for dialogue state generation.
Our experiments show that the schema encoding improves joint accuracy even in few-shot settings.

Even though our approach could perform DST well on target domain with few-shot settings, it required some new data to be fine-tuned.
As part of our future work, we plan to design a DST model for zero-shot settings.

\section*{Acknowledgements}
This work was supported by Institute for Information \&communications Technology Planning \& Evaluation (IITP) grant funded by the Korea government (MSIT) (No. 2021-0-00354, Artificial intelligence technology inferring issues and logically supporting facts from raw text), and by the MSIT (Ministry of Science and ICT), Korea, under the ITRC (Information Technology Research Center) support program (IITP-2022-2020-0-01789) supervised by the IITP (Institute for Information \& Communications Technology Planning \& Evaluation).


\bibliographystyle{acl_natbib}

\clearpage
\appendix
\section{Hyperparameters}
We list the hyperparameters used in our experiments for reproducibility.
\begin{table}[ht]
    \renewcommand{\tabcolsep}{10pt}
    \centering
    \begin{tabular}{l|c}
        \Xhline{5\arrayrulewidth}
        Hidden size & 768 \\
        Embedding size & 768 \\
        Vocabulary size & 30522 \\
        Dropout & 0.3 \\
        Early stopping count & 5 \\
        Max epochs & 40 \\
        Min epochs & 20 \\
        Batch size & 8 \\
        Learning rate & 3e-5 \\
        Gradient clipping & 10 \\
        $\alpha$ & 0.5 \\
        $\beta$ (on SGD) & 3 \\
        $\beta$ (on MultiWOZ) & 5 \\
        \Xhline{5\arrayrulewidth}
    \end{tabular}
    \caption{Hyperparameters used for the experiments in this paper.}
    \label{table:hparams}
\end{table}

\section{User Actions Set}
\label{sec:appendix:action}
\noindent Table \ref{table:action} lists the user actions covered in this paper.
When the user state $U_t$ is generated, each element has three types: (1) action-slot-value triple, (2) action-slot pair, and (3) only action; e.g., (1) \texttt{Inform-restaurant\_location-SanJose}, (2) \texttt{Request-restaurant\_address}, and (3) \texttt{Negate}.
\begin{table}[ht]
    \renewcommand{\tabcolsep}{5pt}
    \centering
    \begin{tabular}{l|c|c}
        \Xhline{5\arrayrulewidth} Action Name & Need Slot & Need Value \\
        \hline Inform & $\checkmark$ & $\checkmark$ \\
        \hline Inform\_Intent* & $\checkmark$ & $\checkmark$ \\
        \hline Request & $\checkmark$ & $\times$ \\
        \hline Request\_Alts & $\times$ & $\times$ \\
        \hline Affirm & $\times$ & $\times$ \\
        \hline Affirm\_Intent* & $\checkmark$ & $\checkmark$ \\
        \hline Select & $\checkmark$ & $\checkmark$ \\
        \hline Negate & $\times$ & $\times$ \\
        \hline Negate\_Intent & $\times$ & $\times$ \\
        \hline Thank\_You & $\times$ & $\times$ \\
        \hline Goodbye & $\times$ & $\times$ \\
        \Xhline{5\arrayrulewidth}
    \end{tabular}
    \caption{List of user actions covered in this paper. * just need an intent as the value, but we added a dummy slot, \texttt{Intent}, to keep the shape of action-slot-value triple; e.g., we used \texttt{Inform\_Intent-Intent-FindRestaurants} instead of \texttt{Inform\_Intent-FindRestaurants} when we make the user state.}
    \label{table:action}
\end{table}

\section{Pre-traing Results}
We evaluated the pre-training performance on KLUE\footnote{\url{https://github.com/KLUE-benchmark/KLUE}.} dataset \cite{park2021klue}, which is a Korean dataset for DST, in addition to SGD.
Table \ref{table:pre:result} shows the pre-training results on SGD and KLUE.
SET-DST outperformed the baselines.
These results demonstrate that SET-DST successfully performs DST with just pre-training.
In the experiment on KLUE, we used KLUE-BERT\footnote{\url{https://huggingface.co/klue/bert-base}.} and KoGPT-2\footnote{\url{https://huggingface.co/skt/kogpt2-base-v2}.} that are large-scale language models pre-trained on Korean corpus.
\begin{table}[ht]
    \renewcommand{\tabcolsep}{5pt}
    \centering
    \begin{tabular}{l|l|l}
        \Xhline{5\arrayrulewidth} \multicolumn{2}{c|}{} & \multicolumn{1}{c}{JA} \\
        \hline \multirow{2}{*}{SGD} & Baseline \small\cite{rastogi2020towards} & 25.40\% \\
        \cline{2-3}  & SET-DST (Ours) & 55.56\% \\
        \hline \multirow{2}{*}{KLUE} & Baseline \cite{park2021klue} & 50.22\% \\
        \cline{2-3}  & SET-DST (Ours) & 57.61\% \\
        \Xhline{5\arrayrulewidth}
    \end{tabular}
    \caption{Pre-training results on SGD and KLUE compared to their baselines.}
    \label{table:pre:result}
\end{table}

\section{Example of dialogues}
We add an example of dialogue with the corresponding user state $U_t$ and dialogue state $D_t$.
Table \ref{table:example} shows an dialogue on three domains: hotels, flights, and attractions.
\begin{table*}[ht]
    \renewcommand{\tabcolsep}{5pt}
    \centering
    \begin{tabular}{ll}
        \Xhline{5\arrayrulewidth} User: & I want to find me a 4 star hotel in Sydney. \\
        \hline \multirow{2}{*}{$U_t$:} & Inform\_Intent - Intent - SearchHotel ; Inform - hotel\_location - Sydney ; \\
        & Inform - star\_rating - 4 \\
        \hline $D_t$: & \{ \textbf{hotel\_location: Sydney}, \textbf{star\_rating: 4} \} \\
        \hline System: & I found 10 option that you may like. A 4 star hotel called 28 Hotel Sydney. \\
        \hline User: & Can you smoke here? Tell me the address. \\
        \hline $U_t$: & Request - smoking\_allowed ; Request - street\_address \\
        \hline $D_t$: & \{ hotel\_location: Sydney, star\_rating: 4 \} \\
        \hline \multirow{2}{*}{System:} & No, smoking is not allowed here. 28 Regent Street, Chippendale New South Wales 2008, \\ 
        & Australia is their address. \\
        \hline \multirow{2}{*}{User:} & Sounds okay. Find me a round trip flight to get there. I will start my trave on day after \\ 
        & tomorrow and I want Premium Economy class tickets. \\
        \hline \multirow{3}{*}{$U_t$:} & Select - place\_name - 28 Hotel Sydney ; Inform\_Intent - Intent - SearchRoundtripFlights ; \\
        & Inform - flights\_departure\_date - day after tomorrow ; Inform - seating\_class - Premium \\ & Economy \\
        \hline \multirow{2}{*}{$D_t$:} & \{ hotel\_location: Sydney, star\_rating: 4, \textbf{place\_name: 28 Hotel Sydney}, \\
        & \textbf{flights\_departure\_date: day after tomorrow}, \textbf{seating\_class: Premium Economy} \} \\
        \hline System: & On which day you will return? \\
        \hline User: & 12th of March is the date on which I will return. \\
        \hline $U_t$: & Inform - return\_date - 12th of March \\
        \hline \multirow{3}{*}{$D_t$:} & \{ hotel\_location: Sydney, star\_rating: 4, place\_name: 28 Hotel Sydney, \\
        & flights\_departure\_date: day after tomorrow, seating\_class: Premium Economy \\
        & \textbf{return\_date: 12th of March} \} \\
        \hline System: & From which city you want to depart? \\
        \hline User: & From Las Vegas. \\
        \hline $U_t$: & Inform - origin\_airport - Lag Vegas \\
        \hline \multirow{3}{*}{$D_t$:} & \{ hotel\_location: Sydney, star\_rating: 4, place\_name: 28 Hotel Sydney, \\
        & flights\_departure\_date: day after tomorrow, seating\_class: Premium Economy \\
        & return\_date: 12th of March, \textbf{origin\_airport: Las Vegas} \} \\
        \hline \multirow{2}{*}{System:} & I have 1 flight for you. You like United Airlines? The onward flight takes off at 4:30 am \\ 
        & and return is take off at 4:55 pm. It has a layover and ticket cost \$697. \\
        \hline User: & Good. Find me some attractions there that have free entry and is child-friendly. \\
        \hline \multirow{2}{*}{$U_t$:} & Inform\_Intent - Intent - FindAttractions ; Inform - free\_entry - True ; \\
        & Inform - good\_for\_kids - True ; Inform - category - dontcare \\
        \hline \multirow{4}{*}{$D_t$:} & \{ hotel\_location: Sydney, star\_rating: 4, place\_name: 28 Hotel Sydney, \\
        & flights\_departure\_date: day after tomorrow, seating\_class: Premium Economy \\
        & return\_date: 12th of March, origin\_airport: Las Vegas, \textbf{free\_entry: True} \\
        & \textbf{good\_for\_kids: True}, \textbf{category: dontcare} \} \\
        \hline System: & You can check out a Sports Venue called ANZ Stadium. \\
        \hline User: & Good. Tell me their phone number. \\
        \hline $U_t$: & Select - attraction\_name - ANZ Stadium ; Request - phone\_number \\
        \hline \multirow{4}{*}{$D_t$:} & \{ hotel\_location: Sydney, star\_rating: 4, place\_name: 28 Hotel Sydney, \\
        & flights\_departure\_date: day after tomorrow, seating\_class: Premium Economy \\
        & return\_date: 12th of March, origin\_airport: Las Vegas, free\_entry: True \\
        & good\_for\_kids: True, category: dontcare, \textbf{attraction\_name: ANZ Stadium} \} \\
        \hline System: & 2 9298 3777 is the phone number. \\
        \hline User: & Great. That's all that I wanted for now. Bye. \\
        \hline $U_t$: & GoodBye \\
        \Xhline{5\arrayrulewidth}
    \end{tabular}
    \caption{Example of dialogue including the user state and dialogue state that we defined in this paper.}
    \label{table:example}
\end{table*}

\end{document}